\documentclass[nonacm,screen]{acmart}
\usepackage{array}
\usepackage{arydshln}
\usepackage{graphicx}
\usepackage{float}
\usepackage[rightcaption]{sidecap}

\AtBeginDocument{%
  \providecommand\BibTeX{{%
    \normalfont B\kern-0.5em{\scshape i\kern-0.25em b}\kern-0.8em\TeX}}}

\setcopyright{mine}
\begin{document}

\title{Time Series Foundation Models and  Deep Learning Architectures for Earthquake Temporal and Spatial Nowcasting}


\author{Alireza Jafari}
% \authornote{All authors contributed equally to this research.}
\email{jrp5td@virginia.edu}
% \orcid{0000-0002-9079-1971}
\affiliation{%
  \institution{University of Virginia, Charlottesville, VA}
  \country{USA}
  \postcode{400740-22904}
}

\author{Geoffrey Fox}
\email{vxj6mb@virginia.edu}
% \orcid{0000-0003-1017-1391}
\affiliation{%
  \institution{University of Virginia, Charlottesville, VA}
  \country{USA}
  \postcode{400740-22904}
}

\author{John B. Rundle}
\email{jbrundle@ucdavis.edu}
% \orcid{0000-0002-1966-4144}
\affiliation{%
  \institution{University of California, Davis, CA and Santa Fe Institute, NM}
  \country{USA}
  \postcode{95616-5270}
}

\author{Andrea Donnellan}
\email{andrea.donnellan@jpl.nasa.gov}
\affiliation{%
  \institution{Jet Propulsion Laboratory California Institute of Technology, Pasadena, CA}
  \country{USA}
}

\author{Lisa Grant Ludwig}
\email{lgrant@uci.edu}
\affiliation{%
  \institution{University	of	California,	Irvine,	CA}
  \country{USA}
}

% \renewcommand{\shortauthors}{Jafari, A and Fox, G}

\begin{abstract}

Advancing the capabilities of earthquake nowcasting, the real-time forecasting of seismic activities remains a crucial and enduring objective aimed at reducing casualties. This multifaceted challenge has recently gained attention within the deep learning domain, facilitated by the availability of extensive, long-term earthquake datasets. Despite significant advancements, existing literature on earthquake nowcasting lacks comprehensive evaluations of pre-trained foundation models and modern deep learning architectures. These architectures, such as transformers or graph neural networks, uniquely focus on different aspects of data, including spatial relationships, temporal patterns, and multi-scale dependencies. This paper addresses the mentioned gap by analyzing different architectures and introducing two innovation approaches called MultiFoundationQuake and GNNCoder. We formulate earthquake nowcasting as a time series forecasting problem for the next 14 days within 0.1-degree spatial bins in Southern California, spanning from 1986 to 2024. Earthquake time series is forecasted as a function of logarithm energy released by quakes. Our comprehensive evaluation employs several key performance metrics, notably Nash–Sutcliffe Efficiency and Mean Squared Error, over time in each spatial region. The results demonstrate that our introduced models outperform other custom architectures by effectively capturing temporal-spatial relationships inherent in seismic data. The performance of existing foundation models varies significantly based on the pre-training datasets, emphasizing the need for careful dataset selection. However, we introduce a new general approach termed MultiFoundationPattern that combines a bespoke pattern with Foundation model results handled as auxiliary streams. In the earthquake case, the resultant MultiFoundationQuake model achieves the best overall performance.

\end{abstract}

\begin{CCSXML}
<ccs2012>
   <concept>
       <concept_id>10010147.10010257</concept_id>
       <concept_desc>Computing methodologies~Machine learning</concept_desc>
       <concept_significance>500</concept_significance>
       </concept>
   <concept>
       <concept_id>10010147.10010178</concept_id>
       <concept_desc>Computing methodologies~Artificial intelligence</concept_desc>
       <concept_significance>300</concept_significance>
       </concept>
    <concept>
        <concept_id>10010147.10010257.10010293.10010294</concept_id>
        <concept_desc>Computing methodologies~Neural networks</concept_desc>
        <concept_significance>300</concept_significance>
        </concept>
    <concept>
        <concept_id>10010147.10010257.10010258.10010262.10010277</concept_id>
        <concept_desc>Computing methodologies~Transfer learning</concept_desc>
        <concept_significance>300</concept_significance>
        </concept>
 </ccs2012>
\end{CCSXML}

\ccsdesc[500]{Computing methodologies~Machine learning}
\ccsdesc[500]{Computing methodologies~Artificial intelligence}
\ccsdesc[300]{Computing methodologies~Transfer learning}
\ccsdesc[500]{Applied computing~Physical sciences and engineering}
\keywords{Earthquake Nowcasting, Deep Learning, Foundation Models, Transformers, Pre-trained models, Graph Neural Networks, Seismic Data.}

% \received{1 June 2025}
% \received[revised]{1 June 2025}
% \received[accepted]{1 June 2025}

\maketitle

\section{Introduction}

Earthquake forecasting represents a dynamic and critical domain of research with profound implications for disaster risk reduction and public safety. The ability to accurately nowcast seismic events and mitigate their impacts is vital for preserving lives and minimizing damage. Forecasting earthquakes is inherently complex due to the absence of consistent and reliable indicators and the infrequent occurrence of major seismic events. This complexity is further compounded by multifaceted long-term and short-term interactions that can transfer between regions \citep{jordan2011operational}. Earthquake nowcasting is a cutting-edge approach in seismology that focuses on real-time nowcasting of seismic activity to assess immediate risk \citep{geohazards3020011}. By analyzing extensive historical and real-time seismic datasets, nowcasting identifies patterns that may signal an impending earthquake. Deep learning, with its robust capacity to process and derive insights from extensive datasets, offers a promising avenue for identifying patterns and potential predictive signals within these data. By leveraging advanced algorithms and computational power, deep learning facilitates a deeper understanding of seismic phenomena, leading to more accurate and reliable earthquake forecasting \citep{rundle2021nowcasting}.

Historically, the study of future earthquakes has relied heavily on statistical models, which operate by identifying patterns to forecast seismic events, as referenced in various studies \citep{de2016statistical, chuang2021development}. However, these conventional methodologies frequently encounter limitations in capturing the complex temporal and spatial patterns inherent in seismic activities \citep{2020EA001097}. Such shortcomings manifest in challenges related to capturing subtle long-term patterns and understanding the interactions between different seismic activities and the diverse nature of seismic sources.  

In recent years, deep learning techniques have emerged as a promising avenue to address these challenges by leveraging the power of neural networks and transformers to learn and extract patterns from vast amounts of seismic data \citep{mousavi2020earthquake, rundle2022nowcasting, geohazards3020011}. A prevalent methodology for earthquake nowcasting involves the utilization of sophisticated neural network architectures, such as Convolutional Neural Networks (CNNs) or Long Short-Term Memory Networks (LSTMs) \citep{perol2018convolutional, harirchian2020earthquake, geohazards3020011}. CNNs generally excel at capturing local patterns within features, while LSTMs are adept at modeling temporal dependencies and long-term trends. 

Graph Neural Networks (GNNs) present a well-suited approach for earthquake forecasting due to their proven ability to capture and model complex relationships within data through graph structures. These models have demonstrated significant promise across various domains, such as stock markets \citep{jafari2022gcnet, jafari2022netpred} and large language models \citep{shariatmadari2024harnessing}, characterized by intricate networked relationships. Surprisingly, despite their potential and success in several domains, the use of GNNs in earthquake forecasting remains relatively underexplored. Only a limited number of studies have adopted GNNs for this purpose, and the graph-based modeling of known effective components, such as fault lines, needs further investigation \citep{graphi1, s22176482, mcbrearty2022earthquake, mcbrearty2023earthquake, graphi2}.

An emerging advancement in deep learning methodologies involves the application of transformer architectures. Transformer models, such as BERT \citep{devlin2018bert} and GPT \citep{radford2018improving}, initially popularized in the domain of natural language processing, have revolutionized how we approach problems by capturing long-range dependencies and understanding context \citep{vaswani2017attention}. However, despite their versatility and success in various applications, transformer models have not been commonly used for earthquake forecasting \citep{sadhukhan2023predicting, saad2022real}. Given their ability to handle sequential data and capture complex temporal patterns, transformers have the potential to significantly enhance earthquake forecasting.

Similarly, pre-trained foundation models have shown powerful capabilities in transferring knowledge across different tasks and domains \citep{raffel2020exploring}. In particular, over the last three years, there have been over sixty projects and 200 references addressing time series foundation models \cite{sciencefmhub}. Pre-trained foundation models have shown considerable promise in time series prediction, effectively capturing temporal dependencies and improving predictive accuracy \citep{liu2024itransformer}. However, despite their success and versatility, these models have not yet been applied to the specific challenge of earthquake forecasting.

In this work, we focus on earthquake time series as they offer a real-time approach to nowcasting seismic activities, which is essential for immediate risk assessment and response \citep{holschneider2014can}. Unlike traditional forecasting methods, nowcasting allows for rapid updates and model predictions, which can significantly enhance preparedness and mitigate the impacts of earthquakes. Furthermore, several physical principles that underlie earthquake time series analysis are discussed in the following references, but we do not delve into them here\citep{jordan2011operational, geohazards3020011, rundle2021nowcasting}. In a later study, we will report on studies of the ETAS-based earthquake simulator \cite{Zhuang2012ETAS,Field2017ETAS, Rundle2023ETAS}, which provides a powerful method to understand the role of different physical effects seen in earthquake time series.

Building on previous research \citep{2022EA002343, 2020EA001097}, we conceptualize earthquake nowcasting as a time series model prediction task. Our objective is to nowcast seismic activities within 0.1-degree spatial bins over a 14-day horizon in Southern California, utilizing data spanning from 1986 to 2024. By focusing on the logarithm of the energy released by earthquakes, we construct a time series for each spatial bin, aiming to accurately capture the intensity and patterns of seismic events. 

\begin{figure}
    \centering
    \includegraphics[width=1.0\linewidth]{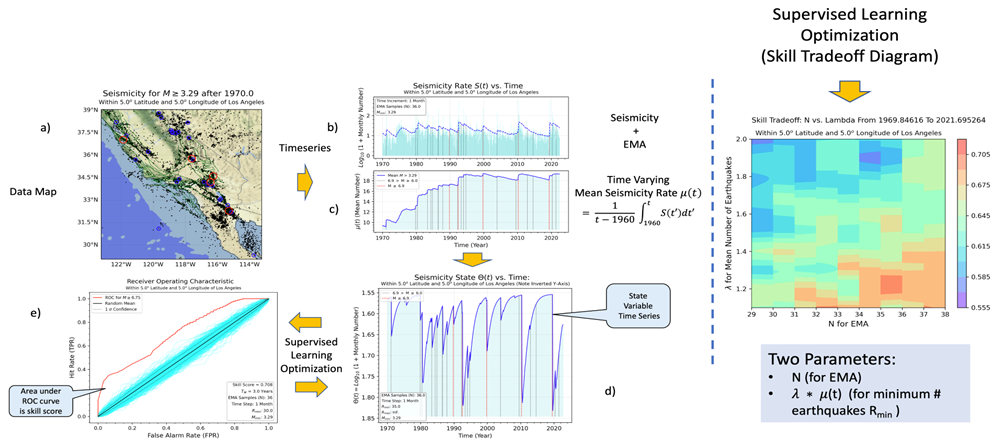}
    \caption{ Illustration of the construction of a nowcast model for California.  The nowcast is a 2-parameter filter on the small earthquake seismicity \cite{Rundle2024QuakeGPT,2022EA002343}.  a)  Seismicity in the Los Angeles region since 1960, M>3.29.  b)  Monthly rate of small earthquakes as cyan vertical bars.  The blue curve is the 36-month exponential moving average (EMA).  c) Mean rate of small earthquakes since 1970.  d)  Nowcast curve that is the result of applying the optimized EMA and corrections for time varying small earthquake rate to the small earthquake seismicity.  e) Optimized Receiver Operating Characteristic (ROC) curve (red line) used in the machine learning algorithm.  Skill is the area under the ROC curve and is used in the optimization.  Skill tradeoff diagram shows the range of models used in the optimization.}
    \label{fig:QuakeGPT}
\end{figure}

Our approach is similar to \cite{geohazards3020011} and \cite{Rundle2024QuakeGPT} whose nowcasting technique is shown in Fig. \ref{fig:QuakeGPT} \citep{2022EA002343}. In this method, an exponential moving average (EMA) is applied to the time series of small earthquakes in southern California, with a correction to account for the relatively poor detection of low-magnitude earthquakes in the time series for the pre-automated era prior to about 1995.  Thus, a 2-parameter filter on the small earthquake time series was constructed and then optimized with machine learning, as shown in Figure 1.  The optimization criterion was the receiver operating characteristic (ROC) skill, which is the area under the ROC curve. We show the value of these physics-motivated observables for Foundation models later in this paper, Table \ref{table:performance_table_2}. 

Building on this nowcasting method, we then developed a time-series transformer model  (QuakeGPT) to predict future values of the time series beyond the observed test data, as shown in Fig. \ref{fig:QuakeGPT}. In this transformer model for time series, we use feature vectors consisting of 36 months of data to predict the next value.  We identify the "keys" with 36 months of data in the training data set, and the corresponding "values" as the value of the next data point.  We identify the "queries" as the 36-month feature vectors in the validation data set that are used to predict the subsequent value of the time series.  

\begin{SCfigure}
    \centering
    \caption{Image showing the application of the trained QuakeGPT transformer to an independent, scaled nowcast validation curve (green shading), followed by prediction of future values beyond the end of the nowcast curve (magenta shading).  In this model, we use 6006 training epochs.  36 previous values are use to predict the next value.  Dots show the predictions and the solid line shows the nowcast curve whose values are to be predicted.  Green dots show the predictions of the transformer up to the last 37 values.  36 blue dots are predictions that were made and then fed back into the transformer to predict the final point (red dot).  In this model, 50 members of an ensemble of runs were used to make the predictions.  The dots represent the mean predictions.  Brown areas represent the 1-sigma standard deviations to the mean values.  In this model 2021 years of simulation data were used to train the model.  The input dimension is 36 points, output dimension is 1 point (36 previous values are used to predict the next value).  Hidden dimension of the neural network in the encoder-decoder layers is 32 neurons.  Number of layers is 2 for this simple model, 1 encoder, 1 decoder.  Number of self-attention heads is 4.  Dropout rate is 0.02, learning rate for the gradient descent Adam optimizer is 0.001.}
    \includegraphics[width=0.6\linewidth]{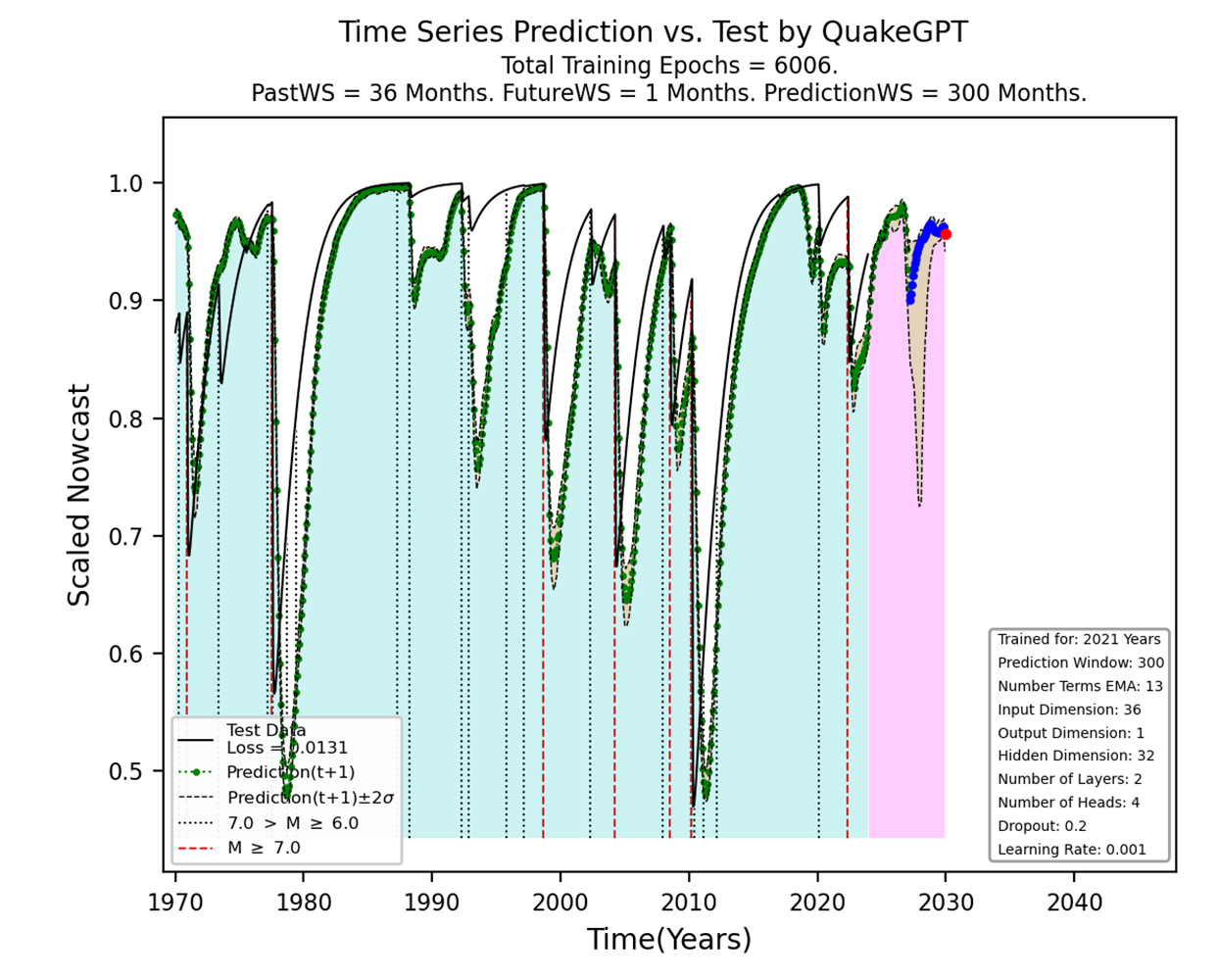}
    
    \label{fig:QuakeGPT2}
\end{SCfigure}

Initial results of a feasibility study are for a similar region studied in this paper, using simulated test data,  and are shown in Fig.  \ref{fig:QuakeGPT2}.  The data here originates from new ERAS (Earthquake Rescaled Aftershock Seismicity) simulated data from \cite{Rundle2024ERAS}.  These results are based on training the transformer model on 2,021 years of ERAS data and then applying it to 53 years of independent test ERAS data (cyan region), a time span similar to the observed data in California.  Then, nowcasts were made for data values beyond the test data by feeding the previously predicted output values into the transformer as inputs to predict future values as described previously.

To address the above-mentioned limitations, we adopt five large pre-trained foundation models for earthquake nowcasting: iTransformer \citep{liu2024itransformer}, PatchTST \citep{nie2023time}, TimeGPT \citep{garza2023timegpt}, Chronos \citep{ansari2024chronos}, and TSMixer \citep{chen2023tsmixer}. The first four models—iTransforemer, PatchTST, TimeGPT, and Chronos— are transformer-based and TSMixer is MLP-based. These models have demonstrated exceptional performance in various domains by effectively handling complex temporal dependencies \citep{wu2021autoformer}. TimeGPT is a generative time-series forecasting model leveraging GPT architecture, pre-trained on a large collection of time series with over 100 billion data points. Chronos, on the other hand, has proven its ability to integrate temporal information by converting time series data into contextual insights, making it particularly useful for our purposes. We pre-train the remaining models on three different datasets to enhance the model's predictive capabilities.

We also modify memory-based models, including DilatedRNN \cite{chang2017dilated}, TFT \cite{lim2021temporal}, and LSTM, to evaluate their suitability for earthquake nowcasting. DilatedRNN has shown strong potential in modeling sequential dependencies over extended periods, while TFT excels in handling multiple time series with varied lengths. LSTM, a widely recognized model in time series forecasting, provides valuable insights into the temporal dynamics of seismic activities.

To explore the capabilities of convolutional layers, we employ models such as TimesNet \cite{bai2018empirical} and TCN \cite{wu2023timesnet}. TimesNet effectively captures local temporal features, while TCN demonstrates robustness in handling long-range dependencies, making both models suitable for the intricacies of earthquake data. Additionally, we incorporate a powerful MLP-based model called TiDE \cite{das2023long}, which showcases impressive performance due to its simplicity and efficiency in capturing non-linear relationships within the data.

We introduce a GNN architecture, called GNNCoder, for earthquake nowcasting that leverages geographical interactions to enhance model prediction accuracy. We create an earthquake graph using the epsilon nearest neighbor algorithm based on the proximity of spatial bins, identifying spatial clustering and patterns. Our model employs Graph Attention Networks (GAT) with an MLP-based encoder-decoder to focus on relevant connections within seismic data, offering a comprehensive framework for accurate and reliable earthquake forecasting.

Furthermore, we introduce the MultiFoundationQuake model, an innovative approach that aggregates multiple foundation models to enhance nowcasting performance. By leveraging the strengths of various pre-trained models, MultiFoundationQuake effectively captures both temporal and spatial dependencies, providing a more robust and accurate forecast of seismic activities. This model highlights the potential of combining diverse pre-trained models to improve earthquake nowcasting.

By advancing the state of the art in earthquake nowcasting, this research significantly contributes to both the deep learning and earthquake research domains. It highlights the critical aspects of earthquake information that must be considered in future forecasting studies. The improved accuracy and reliability of our models have the potential to enhance disaster response efforts, minimize economic losses, and save lives by providing timely and precise nowcasting of seismic events. This research is pivotal in bridging the gap between advanced deep-learning methodologies and practical applications in understanding the probability of earthquake occurrence and mitigation.

In the next section, we will explain the data used in this study, including preprocessing steps and segmentation into spatial bins. The employed methodology and pre-training process are detailed in Section \ref{model_description}, covering model architectures and training procedures. Section \ref{experiment} provides comprehensive information about our experimental setup, evaluation metrics, and baseline comparisons. Finally, we conclude the article in Section \ref{conclusion}, summarizing key findings and suggesting future research directions.

\section{Data}
\label{data}

\begin{figure}
    \centering
    \includegraphics[width=0.6\linewidth]{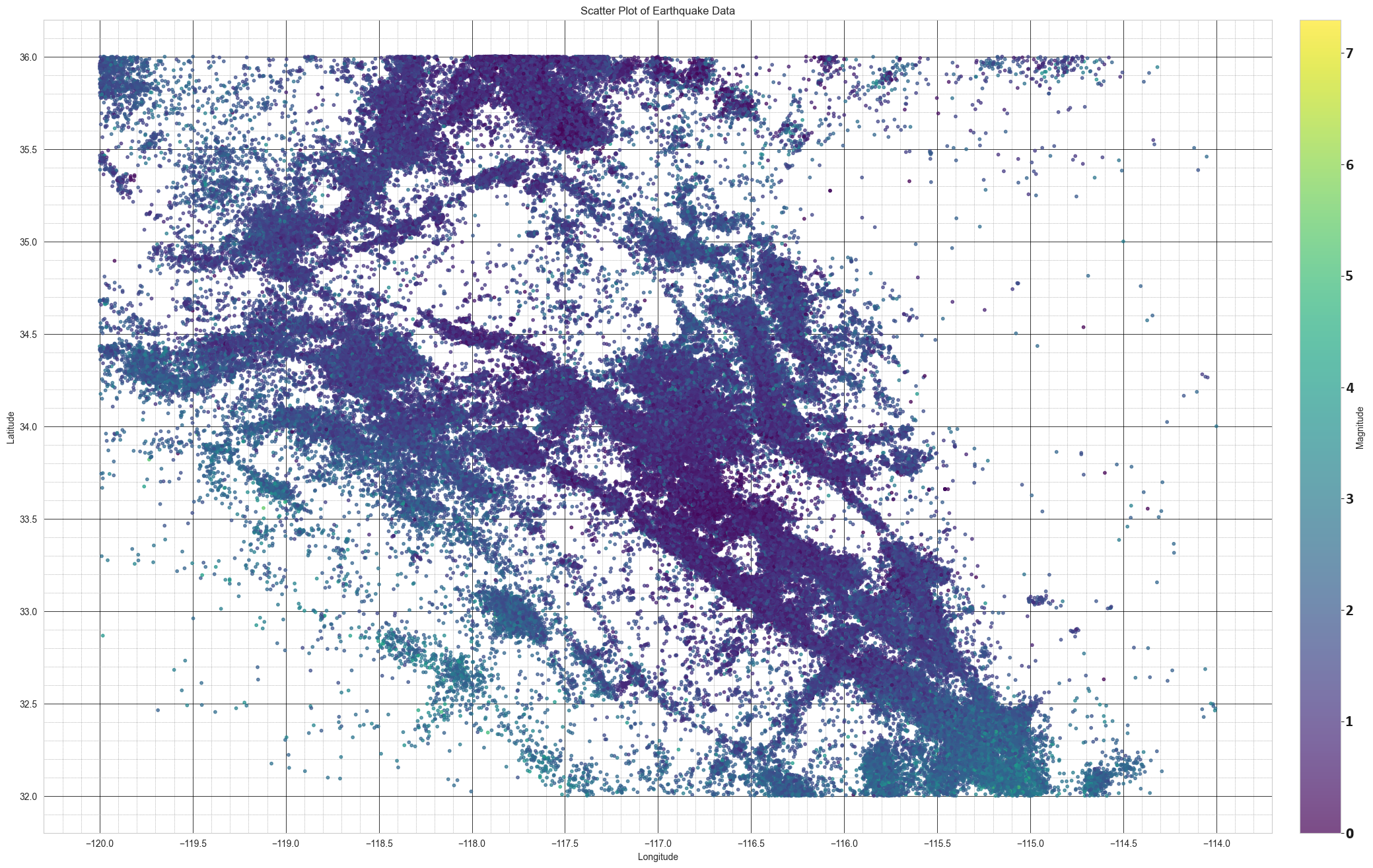}
    \caption{Distribution of earthquake epicenters in Southern California (32°N to 36°N, -120° to -114°) from USGS data (1986-2024). The scatter plot shows the spatial density of seismic events used to analyze and optimize spatial bins for earthquake nowcasting. There is no magnitude cut, with data including all USGS recorded seismic events starting from magnitude 0.}
    \label{fig:scatter_plot}
\end{figure}

In this study, we focus on Southern California, a region renowned for its significant seismic activity and extensive fault lines. The earthquake data utilized were sourced from the US Geological Survey (USGS) online catalogs \citep{USGS_Earthquake_Catalog}. Our analysis encompasses a geographical area defined by a 4-degree latitude span (32°N to 36°N) and a 6-degree longitude range (-120° to -114°), as illustrated in Figure \ref{fig:scatter_plot}. The dataset spans from 1986 to 2024 and includes events recorded with their magnitude, epicenter, depth, and time.

The primary aim of this research is to forecast the cumulative released energy of earthquakes within a two-week horizon, thereby enhancing the understanding and nowcasting of significant earthquakes. To achieve this objective, the designated area of interest is segmented into discrete spatial bins, enabling localized earthquake nowcasting predictions within each bin. Note at this stage, we are exploring \cite{geohazards3020011, Rundle2024QuakeGPT}, both new model architectures, different choices in the quantity nowcast, and the metric of success. As our main interest is understanding different deep-learning architectures, we fix both the nowcasted quantity and metrics across the different time series models investigated here.

\begin{figure}
    \centering
    \includegraphics[width=0.6\linewidth]{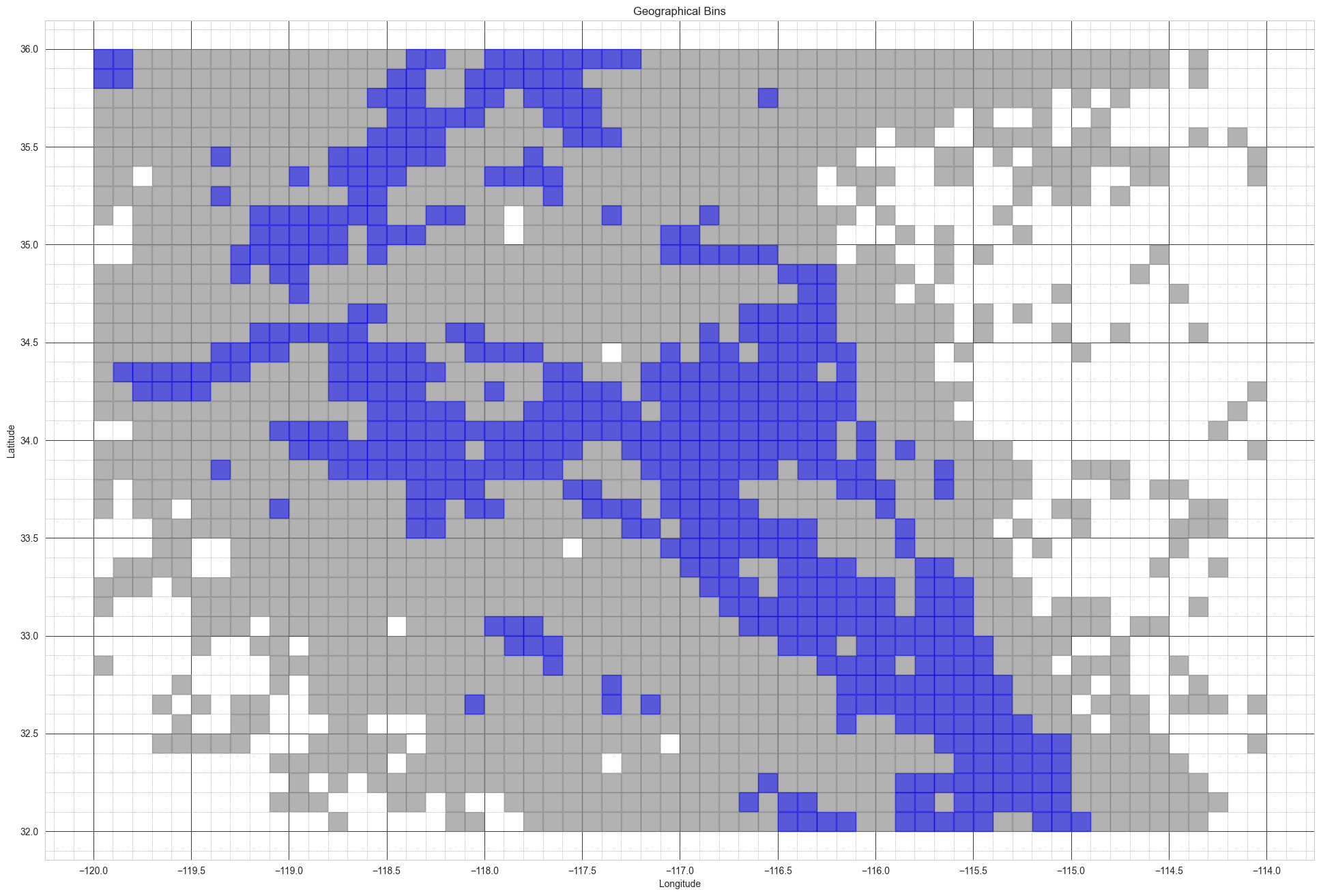}
    \caption{The 500 most active and vulnerable spatial bins, marked in blue, selected for analysis out of the total 2400, based on the frequency of earthquakes from 1986 to 2024. This selection focuses on high-risk areas.}
    \label{fig:most_active_plot}
\end{figure}

Following our previous study \citep{geohazards3020011}, we partition the mentioned area into 0.1x0.1 degree squares, each with a side length of approximately 11 km, resulting in a total of 2400 spatial bins. The decision to use a 0.1-degree grid is inspired by the RELM test initially performed by Ned Field at the USGS, where this specific discretization was applied \citep{Regional_Earthquake2007}.

The quantification of the energy released by earthquakes over a given time period is critical for understanding seismic activity patterns. built on previous studies, we use the following formula to calculate the logarithm of the energy released by seismic events within a specified bin and time period \citep{scholz2019mechanics}:

\begin{equation}
\text{LogEn}_{(bin, time\_period)} = \log(\text{energy}) = \frac{1}{1.5} \log_{10} \left( \sum_{\substack{\text{quakes}}} 10^{1.5 \cdot \text{m}_{\text{quake}}} \right)
\end{equation}

In this formula, \( LogEn_{(bin, time\_period)} \) denotes the logarithm of the total energy released. The summation (\( \sum_{\substack{\text{quakes}}} \)) is carried out over all earthquake events occurring within the designated bin and time period. $m_{quake}$ is the magnitude of an earthquake. Each earthquake's magnitude is raised to the power of 1.5, following the Gutenberg-Richter relationship, which relates earthquake magnitude to energy release \citep{hanks1979moment}. The resulting values are summed to provide a cumulative measure of seismic energy. 

To analyze the temporal distribution of seismic energy, we consider a time window of 14 days. This allows us to create biweekly time period to have samples of seismic activity from 1986 to 2024. By dividing the data into these 14-day intervals, we can systematically assess changes in the total energy release over time. This approach provides a detailed and continuous record of seismic energy, facilitating a better understanding of long-term trends and patterns in earthquake activity.

As previously mentioned, we generate a time series of logarithmic energy for each spatial bin. However, as shown in Figure \ref{fig:scatter_plot}, some bins have experienced relatively few earthquakes over the past 40 years. Forecasting seismic activity in these sparsely active bins would not yield meaningful results. Consequently, we focus our analysis on the 500 most active and vulnerable bins out of the total 2400 within the study area, as illustrated in Figure \ref{fig:most_active_plot}. This selection ensures that we concentrate on areas with a high risk of earthquake occurrence, thereby enhancing the reliability of our nowcasting system.

\begin{figure}
    \centering
    \includegraphics[width=1\linewidth]{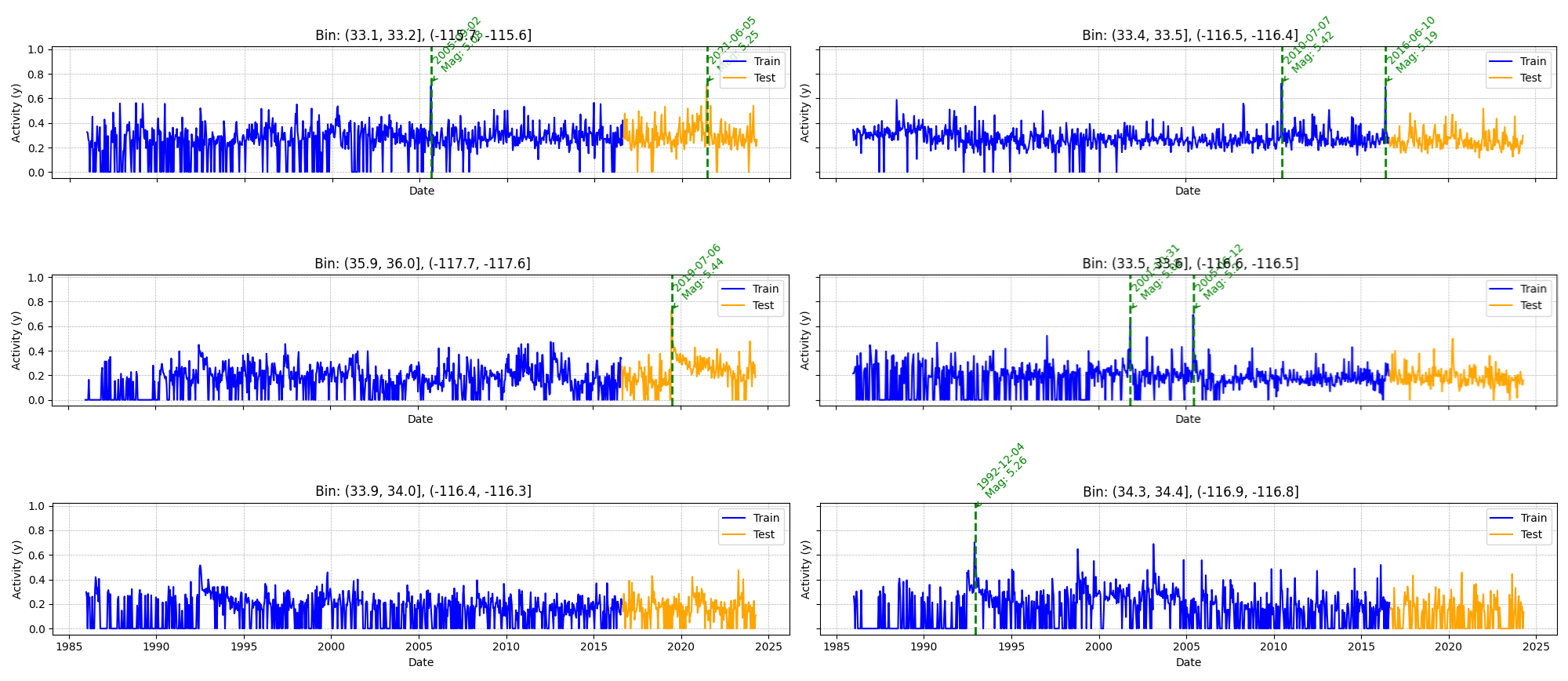}
    \caption{ Six time series from randomly selected spatial bins, highlighting earthquakes of magnitude greater than 5.}
    \label{fig:6_time_series_plot}
\end{figure}

Our dataset spans the years 1986 to 2024, encompassing a total of 1000 two-week samples. We utilize the first 80\% of these samples as training data, while the remaining 20\% are designated as test data. Figure \ref{fig:6_time_series_plot} presents six time series from six randomly selected spatial bins, with earthquakes of magnitude greater than 5 prominently marked in each time series. This plot highlights the temporal distribution and intensity of significant seismic events across different regions.

Note that we must utilize a single feature; however, some of our models can incorporate multiple time series from various spatial bins. To ensure a fair comparison between models, we restrict our input to only time series values, as some models are limited to handling a single feature. However, we generate sequences in each sample consisting of 52 or 130 values to forecast the subsequent value. Additionally, all time series are normalized to have a maximum absolute value of 1 across all spatial and temporal data points. This normalization facilitates consistent and unbiased model evaluation. Although we primarily focus on time series values, we also explore certain scientifically identified features that enhance earthquake nowcasting. In our sub-experiments, we evaluate the performance of models that can accept multiple features using these advanced scientific features \citep{2022EA002343, geohazards3020011}.

\subsection{Graph structure for GNNCoder}
 
In this subsection, we introduce our method for creating the graph structure needed for GNN models. This graph structure enables GNNCoder to effectively utilize and learn from the spatial relationships and interactions between different data points. An earthquake graph structure can be constructed based on various relationships, such as the locations of fault lines and their interactions, or the positions of seismic sensors \citep{graphi2}. However, to objectively evaluate the effectiveness of different deep learning architectures on the multifaceted earthquake problem, we create a geographical graph based solely on the proximity of spatial bins. This approach ensures a fair comparison by excluding additional sources of information and focusing exclusively on the intrinsic spatial relationships within the data.

To construct this graph, we treat each spatial bin as a node and employ the epsilon nearest neighbor graph (epsilon-NNG) algorithm to define the edges \citep{eppstein1997nearest}. The epsilon-NNG algorithm operates by establishing connections between bins based on their proximity to one another. Specifically, spatial bins that fall within a predefined epsilon distance threshold are linked, forming edges in the resulting graph. This method allows for the identification of spatial relationships and dependencies among neighboring bins, providing valuable insights into the spatial clustering and patterns of seismic activity.

\begin{figure}
    \centering
    \includegraphics[width=0.6\linewidth]{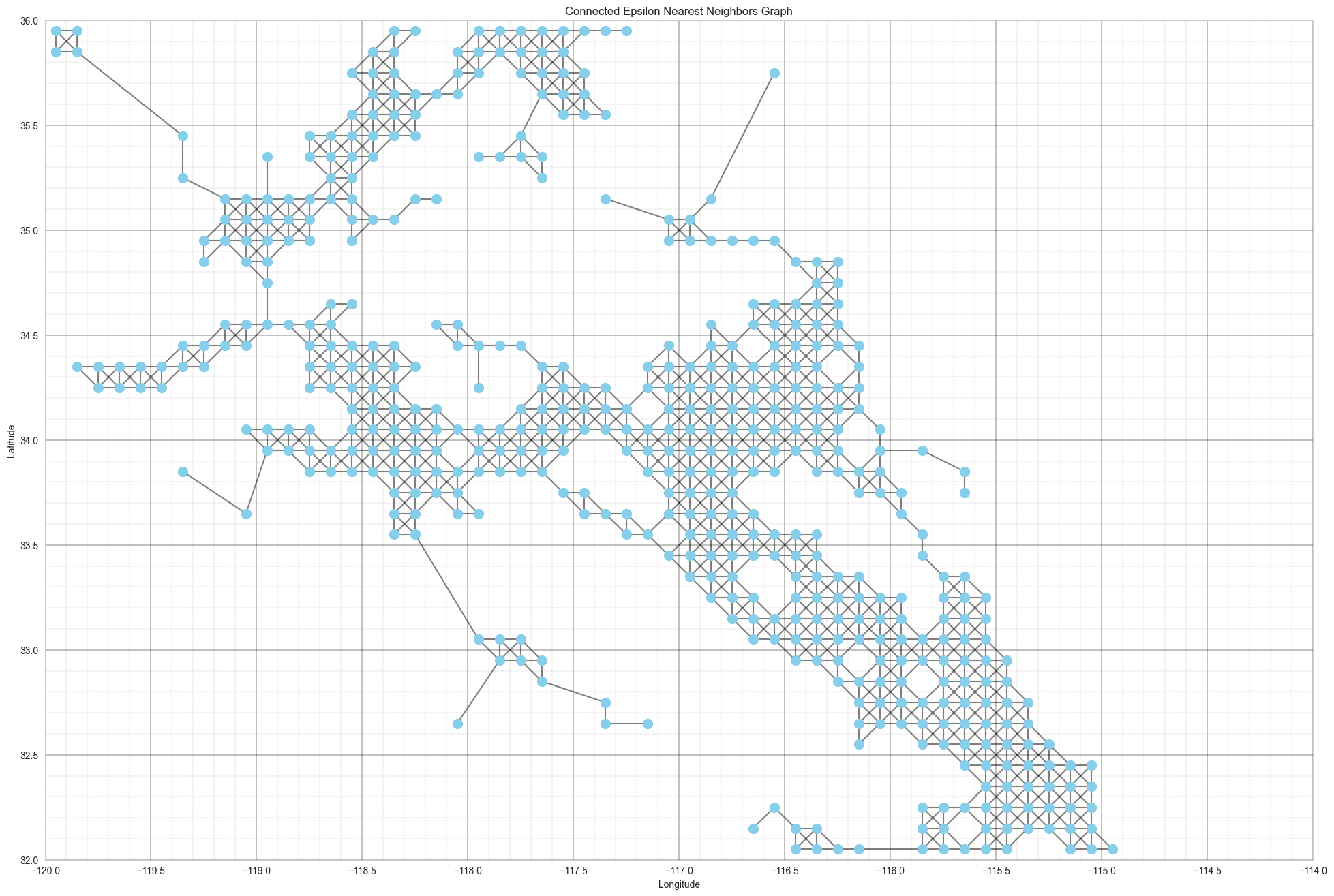}
    \caption{The final graph structure representing the 500 most active bins, created using an epsilon of 0.15 degrees. Initially forming a multi-component graph, components are linked to ensure full connectivity.}
    \label{fig:graph_structure}
\end{figure}

In our experiments, we use an epsilon of 0.15 degrees, ensuring that each bin is connected to its horizontal, vertical, and diagonal neighbors. Focusing on the 500 most active bins, the epsilon-NNG algorithm initially generates a multi-component graph. To achieve full connectivity, we link each component to the nearest node in an adjacent component, thereby creating a single, cohesive graph. This step is essential, as GNN layers aggregate information from specific depths within the graph, and isolated nodes can introduce noise into the embedding process. By utilizing this unweighted graph, we allow the attention mechanism within the GNN layers to determine the significance of each edge, enhancing the model's ability to capture critical spatial relationships. Figure \ref{fig:graph_structure} represents our final graph structure.

\subsection{Pre-training Datasets for Transformer Models}

In this subsection, we describe the pre-training datasets for our selected foundation models. Starting with TimeGPT, it is trained on an extensive dataset comprising time series from various domains such as finance, economics, demographics, healthcare, weather, IoT sensor data, energy, web traffic, sales, transport, and banking. This dataset includes more than 100 billion data points \citep{garza2023timegpt}. 

Chronos is trained using a vast collection of publicly available time series datasets, further enhanced by a synthetic dataset generated via Gaussian processes to improve generalization. The inclusion of synthetic data ensures that Chronos can generalize well, providing robust zero-shot performance on unseen forecasting tasks \citep{ansari2024chronos}.

Additionally, we pre-train iTransformer, PatchTST, and TSMixer on three diverse datasets: the Weather dataset \citep{wu2021autoformer}, the Traffic dataset \citep{lai2018modeling}, and the M4 dataset \citep{makridakis2020m4}. We will discuss each of these datasets in detail in the following subsection. These datasets are selected for their relevance and utility in improving the models' ability to forecast seismic activities. Furthermore, they are well-established datasets commonly used for pre-training on time series problems in numerous studies \citep{wu2021autoformer, haugsdal2023persistence}.

\subsubsection{TrafficL Dataset}
The TrafficL dataset, as presented in \citep{lai2018modeling}, collected by the California Department of Transportation, includes hourly occupancy rates from 862 road sensors, spanning from January 2015 to December 2016. This dataset could be particularly beneficial for earthquake forecasting. Traffic data captures both long-term trends and short-term fluctuations, providing a comprehensive view of temporal dynamics that might be similar to those found in seismic data. 

\subsubsection{Weather Dataset}
The Weather dataset, as presented in \citep{wu2021autoformer}, comprises 21 meteorological measurements recorded every 10 minutes throughout the year 2020 at the Weather Station of the Max Planck Biogeochemistry Institute in Jena, Germany. This dataset might be beneficial in earthquake forecasting as it provides extensive time series data that helps the models learn complex temporal patterns and variability in environmental conditions, which can be similar to the temporal dynamics of seismic activities.

\subsubsection{M4 Dataset}
The M4 dataset, as presented in \citep{makridakis2020m4}, is a comprehensive and large-scale collection of time series data that serves as a standard benchmark for time series forecasting \citep{oreshkin2019n}. It includes various types of time series data, offering a robust baseline for training models to handle different temporal patterns and anomalies. With thousands of time series across multiple domains, the dataset exposes models to a wide range of temporal behaviors, enhancing their adaptability to the unique characteristics of seismic data. As a widely recognized benchmark, the M4 dataset ensures that pre-trained models are calibrated against a high standard, improving their reliability.

We use the monthly dataset from M4, which includes 48,000 time series. The length of these time series varies from 60 to 2812 data points. To ensure sufficient sequence length for creating samples, we sort all the time series in this dataset and retain the 32,000 longest time series in our experiments.

\section{Models Description}
\label{model_description}

This section provides an in-depth description of our comprehensive methodology to develop an earthquake nowcasting system using time series foundation models and advanced deep learning architectures. Our proposed method systematically compares several key architectures to effectively capture and analyze seismic data, highlighting their respective strengths and limitations.

In this work, we introduce two powerful models: MultiFoundationQuake and GNNCoder. MultiFoundationQuake employs a straightforward yet unique approach to aggregate multiple foundation models, significantly enhancing the performance of each individual model.

On the other hand, GNNCoder utilizes a graph attention network, moving beyond traditional methods that predominantly focus on temporal features. This method emphasizes geospatial relationships and clusters, leveraging GNNs to analyze seismic data, innovatively. 

In addition, we modify some powerful foundation models to forecast earthquakes, aiming to analyze their capacities to uncover temporal features as well as indirect spatial features inherent in seismic data. We also incorporate memory-based models that have demonstrated satisfactory results in earthquake forecasting \cite{geohazards3020011, graphi2}.  We discuss each model's structure and specific mechanisms for capturing dependencies in seismic data.

\subsection{Pre-trained Transformer Models}
This section provides a detailed description of the transformer models evaluated in our study for earthquake nowcasting, emphasizing their architectural innovations and unique capabilities.
The pre-trained transformer models evaluated in our study—iTransformer, PatchTST, TimeGPT, and Chronos—bring unique innovations to the task of earthquake nowcasting. By leveraging self-attention mechanisms, these models offer advanced capabilities for capturing complex temporal patterns and enhancing nowcasting accuracy \citep{3295349}. 

%iTransformer
\textbf{iTransformer}, introduced by Liu et al. (2024) \cite{liu2024itransformer}, presents a novel adaptation of the transformer architecture specifically designed for time series forecasting. This innovative model addresses key challenges faced by traditional forecasting methods, such as capturing long-range dependencies and handling multivariate data effectively. Instead of employing a conventional encoder-decoder structure, the iTransformer utilizes an inverted dimension approach. In this configuration, the time points of individual series (each spatial bin) are embedded into variate tokens. Then, the tokens from different regions are processed through multi-head self-attention mechanisms, creating temporal-spatial tokens. This enables the model to capture seismic spatial relationships. 

A critical component of the iTransformer is the attention mechanism, which is pivotal for capturing dependencies in the time series data. The model projects the input data into three distinct vectors: queries (Q), keys (K), and values (V). The attention scores, calculated using the dot product of the query and key vectors, determine the relevance of each element in the sequence relative to others. These scores are scaled and passed through a softmax function to obtain attention weights, which highlight the importance of various elements. The weighted sum of the value vectors, guided by these attention weights, allows the model to aggregate information from the entire sequence from all variates effectively. The model employs multiple attention heads, each independently attending to different parts of the input sequence. 

Then, each variate token undergoes further processing through a position-wise fully connected feed-forward network. This network facilitates the learning of nonlinear representations for each variate token, enhancing the model's ability to represent complex temporal patterns. According to the authors, by leveraging pre-training on the TrafficL and Weather dataset, the model has to be able to gain a robust understanding of temporal patterns, which is crucial for accurate earthquake nowcasting.

%PatchTST
\textbf{PatchTST}, introduced by Nie et al. (2023) \cite{nie2023time}, is a novel model designed to enhance the efficiency and accuracy of multivariate time series forecasting using transformer-based architectures. This model incorporates two key innovations: the segmentation of time series data into subseries-level patches and the concept of channel-independence. By segmenting the time series into patches, PatchTST preserves local semantic information within each patch, which serves as input tokens for the transformer. The input data includes all earthquake time series from each bin (small spatial square), so, It means input tokens for the transformer are representations from a long period of series. This segmentation not only retains important local temporal patterns but also significantly reduces the computational and memory overhead associated with generating attention maps, as the patches are smaller than the full time series. Consequently, PatchTST can process longer historical data more efficiently, enabling it to capture long-term dependencies that are crucial for accurate forecasting.

The second critical component of PatchTST is its channel-independent design. In traditional multivariate time series models, each channel is often treated together, which can lead to complex interactions and increased computational demands. In contrast, PatchTST treats each channel as an independent univariate time series, sharing the same embedding and transformer weights across all channels. The channel-independent design also facilitates the transferability of the model across different datasets. By pre-training on one dataset and fine-tuning on another, PatchTST achieves state-of-the-art forecasting accuracy, showcasing its ability to generalize across different domains \cite{nie2023time}.

%TimeGPT
\textbf{TimeGPT}, introduced by Garza et al. (2023) \cite{garza2023timegpt}, represents a groundbreaking advancement in time series forecasting by developing the first foundation model tailored for this domain. The architecture of TimeGPT leverages insights from transformer-based models. It processes a window of historical values to produce forecasts, incorporating local positional encoding to enrich the input. The architecture follows an encoder-decoder structure with multiple layers, each featuring residual connections and layer normalization. The decoder's output is mapped to the forecasting window dimension through a linear layer, allowing the model to capture the diversity of past events and accurately nowcast potential future distributions.

%Chronos
\textbf{Chronos}, presented by Ansari et al. (2024) \cite{ansari2024chronos}, is a sophisticated text-based framework designed for pre-trained probabilistic time series models, leveraging the principles of natural language processing to enhance time series forecasting. The core innovation in Chronos lies in its tokenization process, where time series values are scaled and quantized into a fixed vocabulary. This approach allows the framework to convert continuous time series data into a discrete sequence of text-like tokens, making it compatible with transformer-based language models. The framework utilizes models based on the T5 family \citep{raffel2020exploring}, with parameter sizes ranging from 20 million to 710 million, to capture varying degrees of complexity in time series patterns. 

The training process for Chronos is both comprehensive and robust. The training corpus is extensive, comprising a large collection of publicly available datasets from diverse domains. To further reinforce the generalization capabilities of the models, the authors generated a synthetic dataset using Gaussian processes. This combination of real-world and synthetic data ensures that Chronos models can learn intricate temporal dependencies and variances, preparing them for a wide array of forecasting scenarios.

\subsection{Graph Neural Networks models}

We propose an innovative approach, called GNNCoder, that enhances the understanding of spatial relationships among geological regions using Graph Neural Networks and enCoder-deCoder components. This methodology creates a holistic framework for earthquake forecasting, addressing the limitations of existing models that often overlook complex interactions between geological features and the temporal evolution of seismic activities.

Upon our introduced earthquake-correlated graph structure, GNNCoder incorporates the roles of faults and other geographical entities. GNNs are particularly adept at modeling and analyzing complex dependencies in graph-structured data, making them ideal for capturing the intricate geographical interactions and dependencies inherent in earthquake data.

Our model architecture includes an MLP-based encoder-decoder component, complemented by multiple Graph Attention Network (GAT) layers. The encoder-decoder, constructed with dense layers, is essential for capturing and transforming input data into a meaningful representation. The GAT layers enhance the model's capacity to handle graph-structured data by dynamically adjusting the importance of neighboring nodes through an attention mechanism \cite{velickovic2018graph}. This capability is crucial for earthquake forecasting, as it enables the model to focus on the most relevant connections and interactions within the seismic data. Ultimately, the model architecture concludes with an additional dense layer designed to forecast seismic energy values. 

The attention mechanism allows GATs to adaptively focus on specific parts of the graph relevant to the task. A key feature of GATs is their utilization of self-attention, or intra-graph attention mechanisms, to compute the hidden representations of each node in the graph. The attention coefficients, which are central to this process, are computed as follows:
\begin{equation}
    \alpha_{ij} = \frac{\exp(\text{LeakyReLU}(\mathbf{a}^T[\mathbf{W}\mathbf{h}_i \| \mathbf{W}\mathbf{h}_j]))}{\sum_{k \in \mathcal{N}_i} \exp(\text{LeakyReLU}(\mathbf{a}^T[\mathbf{W}\mathbf{h}_i \| \mathbf{W}\mathbf{h}_k]))}
\end{equation}
where $\mathbf{h}_i$ is the feature vector of node $i$, $\mathbf{W}$ is a weight matrix, $\mathbf{a}$ is a weight vector in the attention mechanism, $\|$ denotes concatenation, and $\alpha_{ij}$ is the attention coefficient between nodes $i$ and $j$. Unlike other graph neural network models that depend on the entire neighborhood's aggregate information, GATs allow for the weighting of nodes’ features based on their relevance \cite{velickovic2018graph, wang2019heterogeneous}.

In our experiments, we utilize three different GNNCoder, varying from 1-layer to 3-layer GAT architectures. The 1-layer GAT model consists of a single graph attention layer that aggregates information from the immediate neighbors of each node using attention scores to weigh the importance of these neighbors. Each node's features are combined with their neighbors' features, weighted by the attention scores, followed by a non-linear activation function. The 2-layer and 3-layer GAT models extend this architecture by adding more graph attention layers. These additional layers allow the model to capture dependencies up to two and three hops away, respectively, thereby enhancing its ability to learn broader spatial interactions. However, increasing the depth also increases computational complexity and the risk of overfitting, as highlighted by \cite{velickovic2018graph}.

\subsection{Memory-Based Models}

Earthquake nowcasting relies heavily on the analysis of temporal patterns in seismic data. Aftershocks and foreshocks exhibit patterns that are critical for understanding and forecasting seismic activity \citep{2020EA001097}. Seismic data is inherently complex, characterized by irregular intervals and varying magnitudes of seismic waves. Memory-based models are adept at identifying and learning from these patterns due to their ability to maintain long-term dependencies. We use memory-based models such as DilatedRNN, TFT, and LSTM to forecast earthquake time series, and we provide a detailed description of them below.

%DilatedRNN
\textbf{DilatedRNN}, as described by Chang et al. (2017) \cite{chang2017dilated}, is a novel architecture designed to address the challenges of learning long-term dependencies in sequential data. Traditional RNNs, such as LSTMs and GRUs, often struggle with these dependencies due to issues like vanishing and exploding gradients. DilatedRNN introduces a dilation mechanism inspired by dilated convolutions used in CNNs. This mechanism allows the model to skip certain time steps and capture longer temporal dependencies more efficiently. By incorporating dilations into the recurrent architecture, DilatedRNN can effectively balance between capturing short-term and long-term dependencies without a significant increase in computational complexity.

The architecture of DilatedRNN is characterized by its unique dilation patterns, which specify the intervals at which the model accesses past time steps. These patterns are carefully designed to ensure that the model captures a wide range of temporal dependencies. For instance, a dilation pattern might access every other time step in the first layer, every fourth time step in the second layer, and so on. This hierarchical approach allows the network to maintain a larger receptive field and efficiently integrate information from various time scales. The use of dilations mitigates the problem of long-term dependency learning by reducing the effective path length through the network, which in turn enhances gradient flow and model performance. 

%TFT
\textbf{Temporal Fusion Transformer (TFT)}, detailed by Lim et al. (2021) \cite{lim2021temporal}, is a sophisticated attention-based architecture designed for interpretable multi-horizon time series forecasting. TFT integrates the capabilities of deep learning with the necessity for interpretability, leveraging both recurrent layers for local sequence processing and self-attention mechanisms to capture long-term dependencies. This dual approach enables TFT to learn temporal relationships at multiple scales effectively.

A key feature of the TFT is its specialized components that ensure the model's high performance and interpretability. These include variable selection networks that judiciously choose relevant features from a potentially large set of inputs, thereby enhancing the model's ability to focus on the most informative variables. Additionally, TFT employs gating layers that dynamically suppress irrelevant or redundant components within the model, which not only streamlines the computational process but also mitigates overfitting. By combining these elements, TFT achieves a robust balance between complexity and interpretability, offering insights into how different variables and temporal patterns influence the forecasting outcomes.

%The Long Short-Term Memory (LSTM) network, introduced by Hochreiter and Schmidhuber (1997) \cite{hochreiter1997long}, is well-known for its capability to learn long-term dependencies. It consists of a series of LSTM cells, each with input, output, and forget gates that regulate the flow of information. This gating mechanism allows the LSTM to retain relevant information over long sequences while discarding irrelevant data, making it highly effective for time series prediction.

\subsection{Convolutional and MLP-Based Models}

We also employ several powerful and renowned convolutional models due to their ability to automatically and adaptively learn spatial hierarchies from seismic patterns. Furthermore, we use MLP-based models that excel at integrating diverse data sources, such as historical earthquake records.

\textbf{TSMixer}, as described by Chen et al. (2023) \cite{chen2023tsmixer}, represents a novel approach to time series forecasting that leverages the simplicity and effectiveness of multi-layer perceptrons. Unlike transformers and memory-based models, which often rely on recurrent or attention-based mechanisms to capture temporal dependencies, TSMixer utilizes mixing operations along both the time and feature dimensions. This design choice allows the model to efficiently extract information and capture complex dynamics inherent in multivariate time series data. By focusing on linear models, TSMixer demonstrates that high-capacity architectures are not always necessary for achieving state-of-the-art performance, challenging the prevailing notion that more complex models are inherently superior.

\textbf{TimesNet}, described by Wu et al. (2023) \cite{wu2023timesnet}, is an innovative framework designed to address the inherent challenges in time series analysis by transforming 1D temporal data into 2D representations. The core idea behind TimesNet is the decomposition of complex temporal variations into intraperiod and interperiod variations, which are then mapped into 2D tensors. This transformation is pivotal as it allows the model to utilize 2D convolutional kernels to effectively capture and process temporal patterns that are otherwise difficult to discern in a 1D format. By embedding intraperiod variations into the columns and interperiod variations into the rows of the 2D tensors, TimesNet leverages the power of 2D convolutional networks to achieve superior representation and modeling of time series data.

The primary component of TimesNet is the TimesBlock, a task-general backbone specifically designed for time series analysis. TimesBlock is equipped with a parameter-efficient inception block that adapts to the multi-periodicity inherent in time series data. This block can dynamically discover and extract intricate temporal variations from the transformed 2D tensors. The inception mechanism within TimesBlock enables the model to handle multiple scales of temporal patterns, thereby enhancing its ability to forecast, classify, impute, and detect anomalies across various time series datasets. The adaptability and efficiency of TimesBlock make it a versatile tool for a wide range of time series analysis tasks.

\textbf{Temporal Convolutional Network (TCN)}, introduced by Bai et al. (2018) \cite{bai2018empirical}, is an innovative architecture that leverages convolutional layers for sequence modeling tasks, challenging the traditional dominance of recurrent networks like LSTMs. TCNs are designed to handle sequence data by applying convolutional layers across the temporal dimension, thus capturing temporal dependencies through a hierarchy of filters. This approach enables TCNs to model long-range dependencies more effectively than recurrent networks, as they avoid the vanishing gradient problem typically associated with deep recurrent architectures. A key characteristic of the TCN is its use of dilated convolutions. Dilated convolutions allow the network to have a larger receptive field without significantly increasing the number of parameters or the computational cost. This dilation means that the convolutional layers can skip certain inputs, allowing the TCN to aggregate information from a wider range of previous time steps, making it particularly well-suited for tasks requiring long memory.

\textbf{Time-series Dense Encoder (TiDE)}, described by Daset al. (2023) \cite{das2023long}, is an innovative approach designed to address the challenges of long-term time-series forecasting. Unlike traditional models that either rely on the simplicity of linear methods or the complexity of Transformer-based architectures, TiDE leverages the strengths of an MLP-based encoder-decoder framework. This model offers a balanced combination of simplicity, speed, and the capability to handle both covariates and non-linear dependencies in the data. At its core, TiDE uses a dense encoding mechanism that efficiently captures the temporal dependencies inherent in time-series data, ensuring accurate and robust forecasts over extended horizons. The architecture of TiDE is built around an encoder-decoder structure, where the encoder processes input time-series data to generate a dense representation, and the decoder utilizes this representation to produce the forecast. This structure allows TiDE to effectively learn from historical data while adapting to the presence of external variables (covariates), which can influence the future trajectory of the time series.

\subsection{MultiFoundationQuake}

Finally, we introduce MultiFoundationQuake, an innovative approach for earthquake forecasting that leverages the power of multiple foundation models to enhance the accuracy and robustness of nowcasting. The architecture of MultiFoundationQuake consists of two main components: foundation models and a pattern model. The foundation models, such as iTransformer, TFT, and PatchTST, each bring unique strengths to the table. For example, the iTransformer utilizes a specific transformer architecture for capturing long-range dependencies, whereas other models might excel in different areas. The outputs from these foundation models serve as inputs to a pattern network. A pattern network is a non-foundation model trained directly on the target task, focusing solely on learning patterns relevant to that specific domain. This concept can be extended to develop a MultiFoundationPattern model for other domains.

In MultiFoundationQuake1, we utilize six models, including DilatedRNN, iTransformer-TrafficL, TFT, TCN, and TSMixer-TrafficL, in the first component, and adopt an LSTM model for the pattern model in the second component. The LSTM is adept at learning sequential dependencies and temporal dynamics, making it well-suited for processing the enriched feature representations provided by the foundation models.

In MultiFoundationQuake2, we follow a similar architecture but utilize a Graph Attention Network (GAT) for the pattern model in the second component. The choice of GAT is driven by its ability to capture temporal dependencies across all locations, while also considering temporal patterns between neighboring areas. This approach is crucial for effectively modeling the interactions between different seismic regions, enhancing the overall accuracy of the earthquake nowcasting process.

The training process of MultiFoundationQuake involves several key steps. Initially, each foundation model is individually pre-trained on a pre-training dataset based on its designed purpose. In addition to the foundation models, our approach can incorporate other large models as well. Each foundation model is subsequently fine-tuned on seismic data, encapsulating various temporal patterns and dependencies relevant to earthquake nowcasting. These features are concatenated and fed into the pattern model, which is trained to learn the sequential dependencies and improve the accuracy of earthquake nowcasting.

MultiFoundationQuake offers several advantages over traditional earthquake nowcasting models. By combining the strengths of multiple foundation models, it captures a wider range of temporal patterns and dependencies. The integration of diverse feature representations ensures that the model is robust to various seismic data characteristics, leading to more accurate and reliable earthquake nowcasting. Additionally, the modular design of MultiFoundationQuake allows for the incorporation of additional foundation models in the future, enhancing its scalability and adaptability to evolving seismic nowcasting techniques.

\subsection{Model Training and Implementation}

Prior to training, the raw earthquake data is preprocessed to generate time series inputs for each spatial bin. The dataset is divided into 14-day intervals to create biweekly samples of seismic activity. The energy released by earthquakes within each bin and time period is calculated using the logarithm of the summed seismic energy. This preprocessing step ensures that the input data is normalized, facilitating the training process.

In our experiments, we utilize transfer learning by performing both pre-training and fine-tuning through supervised learning. The input sequences (X) consist of either 52 or 130 values representing past seismic activities, with the subsequent value serving as the target (Y). This means that 2 or 5 years of information is used to forecast the subsequent value, enabling the models to learn temporal dependencies effectively. 

The training process involves optimizing the model parameters to minimize the model prediction error, measured using the Mean Squared Error (MSE) loss function. MSE is chosen for its ability to emphasize larger errors, which is crucial for capturing significant seismic events. The training is conducted over multiple epochs. Hyperparameter tuning is performed to optimize model performance. Key hyperparameters include the learning rate, batch size, and the number of layers and attention heads for transformer models.

For the following models (iTransformer, PatchTST, TSMixer), transfer learning is applied by pre-training on large-scale datasets such as Weather, TrafficL, and M4. This step enables the models to learn general temporal patterns and improve their performance on the earthquake nowcasting task. The pre-trained models are then fine-tuned on the earthquake dataset, allowing them to adapt to the specific characteristics of seismic data. The Chronos and TimeGPT models are already trained on large-scale datasets and, according to the authors, do not require fine-tuning.

For the transformer models, we utilize the Python package \citep{olivares2022library_neuralforecast}, which is based on PyTorch. For the GNN models, we employ the Python package \citep{grattarola2020graph}, which is built on TensorFlow.

This study is based on a variant of the Earthquake code in the  MLCommons \citep{mlcommons_homepage} Science benchmarks \citep{von_laszewski2023, thiyagalingam2023, mlcommons_science2023, mlcommons_github}. We will submit our measurements there as our answer to their challenge to improve scientific discovery in this area.

\section{Models Evaluation and Comparison}
\label{experiment}

In this section, we detail our experimental setup and evaluation process and present results and discussions. We compare the performance of advanced deep learning architectures and foundation models on earthquake datasets. Furthermore, we conduct deep investigations of earthquake time series, spatial dependencies, and feature analysis to provide comprehensive insights into model performance. We begin by explaining the evaluation metrics, followed by a comprehensive presentation of the results for both the proposed models and the baselines.

\subsection{Evaluation Metrics}

The scientific objective of the present work is to enhance the quality of earthquake nowcasting using deep learning in a region of Southern California. Similar to that used in previous work \cite{rundle2022nowcasting, rundle2021nowcasting}, we use Normalized Nash-Sutcliffe Efficiency (NNSE) to evaluate the models \citep{nossent2012application}. NNSE is a normalized statistic that determines the relative magnitude of the residual variance compared to the measured data variance. It is used to assess the predictive power of earthquake nowcasting models. The formula for NSE is:

\begin{align}
NSE = 1 - \frac{\sum_{i=1}^{n} (O_i - P_i)^2}{\sum_{i=1}^{n} (O_i - \bar{O})^2} \\
NNSE = 1 / (2-NSE)
\end{align}

where, \( O_i \) is the observed value at time \( i \), \( P_i \) is the nowcasted value at time \( i \), \( \bar{O} \) is the mean of the observed values, \( n \) is the number of observations.

 This metric ranges from 0 to 1, where 1 signifies a perfect match between the model predictions and the observations, and 0.5 indicates that the model's predictions are as accurate as the mean of the observed data. This metric is particularly useful in the context of earthquake nowcasting as it provides a clear measure of how well the model's predictions match the observed data, accounting for the variability inherent in earthquake occurrences \citep{geohazards3020011}.

To comprehensively evaluate the performance of our models, we also employ two other metrics: Mean Squared Error (MSE), and Mean Absolute Error (MAE). These metrics offer a thorough understanding of various facets of the models' performance. These metrics provide a detailed understanding of various aspects of model performance. However, NNSE holds a significant advantage over MSE and MAE, as it is not dataset-dependent. This allows for a more consistent comparison of model results, thereby enhancing the reliability and accuracy of seismic nowcasting in future applications.

\subsection{Results and discussion}

The performance evaluation of various deep learning models for earthquake nowcasting in Southern California reveals significant insights into the strengths and weaknesses of each approach. Table \ref{table:performance_table} defines the model characteristics and provides a detailed comparison of the models based on key metrics such as NNSE, MSE, and MAE. The third column of the table indicates a model is Foundation model (F) or Pattern model (P). F refers to models that are pre-trained on large, diverse datasets to capture general temporal patterns, which can be then fine-tuned for specific tasks like earthquake nowcasting. On the other hand, P models are trained directly on the target task, focusing solely on learning the patterns relevant to that specific domain. The table is sorted by decreasing MSE. The results highlight the importance of model architecture and pre-training datasets in enhancing nowcasting accuracy for seismic activities.

\begin{table}[ht]
\centering
\setlength{\tabcolsep}{7pt}
\begin{tabular}{l|c|c|c|c|c|c|c}
\hline
Model & Architecture & Type & Pre-training & Fine-tuning & MSE & MAE & NNSE \\
\hline
TimeGPT & Transformer & F & A broad dataset & None & 0.01042 & 0.0593 & 0.5484 \\
iTransformer-M4 & Transformer & F & M4 & Earthquake & 0.00702 & 0.0537 & 0.5902 \\
TSMixer-M4 & MLP & F & M4 & Earthquake & 0.00651 & 0.0535 & 0.6081 \\
Chronos & Transformer & F & A broad dataset & None & 0.00650 & 0.0519 & 0.6087 \\
PatchTST-TrafficL & Transformer & F & TrafficL & Earthquake & 0.00644 & 0.0501 & 0.6107 \\
TiDE & MLP & P & None & Earthquake & 0.00643 & 0.0519 & 0.6110 \\
TSMixer-TrafficL & MLP & F & TrafficL & Earthquake & 0.00643 & 0.0505 & 0.6111 \\
TimesNet & CNN & P & None & Earthquake & 0.00643 & 0.0560 & 0.6112 \\
PatchTST-M4 & Transformer & F & M4 & Earthquake & 0.00641 & 0.0504 & 0.6117 \\
PatchTST-Weather & Transformer & F & Weather & Earthquake & 0.00641 & 0.0502 & 0.6119 \\
iTransformer-TrafficL & Transformer & F & TrafficL & Earthquake & 0.00639 & 0.0513 & 0.6125 \\
TCN & CNN & P & None & Earthquake & 0.00637 & 0.0535 & 0.6132 \\
VanillaTransformer & Transformer & P & None & Earthquake & 0.00635 & 0.0498 & 0.6141 \\
TFT & \small{Transformer+RNN} & P & None & Earthquake & 0.00635 & 0.0555 & 0.6142 \\
GNNCoder 2-layer & GNN & P & None & Earthquake & 0.00632 & 0.0520 & 0.6153 \\
LSTM & RNN & P & None & Earthquake & 0.00631 & 0.0514 & 0.6156 \\
DilatedRNN & RNN & P & None & Earthquake & 0.00630 & 0.0510 & 0.6159 \\
GNNCoder 3-layer & GNN & P & None & Earthquake & 0.00629 & 0.0524 & 0.6162 \\
GNNCoder 1-layer & GNN & P & None & Earthquake & 0.00628 & 0.0522 & 0.6166 \\
MultiFoundationQuake1 & Hybrid+LSTM & F+P & Several datasets & \small{Earthquake+FMs} & 0.00626 & 0.0516 & 0.6174 \\
MultiFoundationQuake2 & Hybrid+GNN & F+P & Several datasets & \small{Earthquake+FMs} & 0.00625 & 0.0514 & 0.6175 \\
\hline
\end{tabular}
\caption{Comparison of the performance of deep learning models for earthquake nowcasting in Southern California, ranked by MSE in descending order. The table compares various models used in this work, detailing their architectures, types (F for Foundation Model, P for Pattern Model), and datasets for pre-training and fine-tuning. In case of Patterns, there is no pre-training, and fine-tuning is a supervised training.}
\label{table:performance_table}
\end{table}

Our introduced models, MultiFoundationQuake1, MultiFoundationQuake2, and GNNCoder, demonstrated superior performance across multiple metrics. MultiFoundationQuake2, as the best model, achieved an MSE of 0.00625, an MAE of 0.0514, and an NNSE of 0.6175. This model leverages a hybrid architecture that combines several foundation models with a GNN as the pattern model. The GNN's ability to capture both spatial dependencies and temporal patterns across different seismic regions resulted in improved performance.

MultiFoundationQuake1 follows a similar foundation model structure but replaces the GNN with an LSTM for the pattern model. MultiFoundationQuake1 achieved an MSE of 0.00626, an MAE of 0.0516, and an NNSE of 0.6174. This demonstrates the effectiveness of LSTM in capturing sequential dependencies and temporal dynamics, although MultiFoundationQuake2's GNN slightly outperformed it by better leveraging spatial relationships.

The GNNCoder models, particularly the one-layer version, also showed strong performance. The GNNCoder 1-layer achieved an MSE of 0.00628, an MAE of 0.0522, and an NNSE of 0.6166. These results suggest that the one-layer GNNCoder effectively captures the spatial relationships inherent in seismic data, leveraging the proximity of spatial bins to nowcast earthquake activities accurately. The slightly lower NNSE values in the three-layer and two-layer GNNCoders (0.6162 and 0.6153, respectively) indicate that increasing the network depth can enhance performance, but may also introduce additional complexity, which we discuss in subsection \ref{spatial_analysis}.

The DilatedRNN model also performed well, with an NNSE of 0.6159 and an MSE of 0.00630, indicating its capability to model temporal dependencies effectively. The LSTM model, known for its effectiveness in time series forecasting, showed comparable performance to the DilatedRNN, with an MSE of 0.00631, an MAE of 0.0514, and an NNSE of 0.6156. However, their performances were slightly less favorable compared to GNNCoder. This suggests that while the DilatedRNN and LSTM capture sequential patterns efficiently, they may not fully exploit the spatial relationships between seismic events as effectively as the GNNCoder.

Pre-trained foundation models, including iTransformer, TimeGPT, PatchTST, and TSMixer, demonstrated varying degrees of success, heavily influenced by their pre-training datasets. TimeGPT, as the first foundation model in this domain, performed poorly with an MSE of 0.01042 and the lowest NNSE of 0.5484. This poor performance highlights a significant issue: pre-training on a huge dataset is not necessarily sufficient for achieving high accuracy in earthquake nowcasting. Additionally, the iTransformer-M4 model, with an MSE of 0.00702 and an NNSE of 0.5902, emphasizes that pre-training on an irrelevant dataset can considerably decrease the accuracy of earthquake nowcasting. Conversely, the iTransformer-TrafficL model achieved an NNSE of 0.6125 and an MSE of 0.00639, suggesting that pre-training on the TrafficL dataset, which captures temporal dynamics from a road network, provided beneficial insights for earthquake nowcasting. 

The PatchTST-Weather model, with an MSE of 0.00641 and an MAE of 0.0502, showed that weather data can offer valuable pre-training information, enhancing the model's ability to capture complex temporal patterns in seismic data. Similarly, the PatchTST-M4 and PatchTST-TrafficL models showed moderate performance, with MSEs of 0.00641 and 0.00644, respectively. These results underscore the critical role of selecting appropriate pre-training datasets to improve transformer model performance in specific domains. However, the VanillaTransformer outperformed all pre-trained models, demonstrating that focusing solely on learning patterns relevant to specific domain data (in our case, earthquake data) is more important for achieving high predictive accuracy in nowcasting.

The Chronos, a textual model, demonstrated moderate performance with an MSE of 0.00650 and an NNSE of 0.60875. This suggests that converting time series values to text-like tokens struggles with handling complex temporal dependencies, especially in earthquake data. Additionally, the results indicate that integrating spatial information might improve its nowcasting accuracy.

The TimesNet model, designed to capture local temporal features, and the TiDE model, which focuses on non-linear relationships, both showed moderate performance. TimesNet achieved an MSE of 0.00643 and an NNSE of 0.6112, while TiDE had an MSE of 0.00643 and an NNSE of 0.6110. These models, while effective, did not match the performance of the best GNN and RNN models, indicating that capturing spatial and temporal dependencies is crucial for accurate earthquake nowcasting.

From a broader perspective, pre-trained FMs exhibited weaker performance compared to pattern models for two key reasons. First, the higher MSE and lower NNSE scores among FMs highlight the challenges these models face in transferring learned knowledge from broad or irrelevant pre-training datasets to the specific task of earthquake nowcasting. In contrast, pattern models, which focus on direct learning from earthquake data, consistently achieved lower MSEs and higher NNSEs, demonstrating their effectiveness in this domain. 

Second, GNN models excel in understanding spatial relationships by performing specialized graph-based analyses that aggregate data from neighboring regions. In contrast, FMs like iTransformer attempt to derive spatial information from the entire dataset, including regions that may be irrelevant to a target area. This broad approach can introduce noise and hinder the model's ability to distinguish meaningful patterns and dependencies.

\subsubsection{Earthquake Time Series Analysis}

Earthquake time series analysis involves the investigation of temporal patterns within seismic activity data to forecast significant peaks, indicative of substantial energy release during large seismic events. This approach is vital for understanding and nowcasting the occurrence of earthquakes over time, enabling better preparedness and response strategies.

\begin{figure}
    \centering
    \includegraphics[width=1\linewidth]{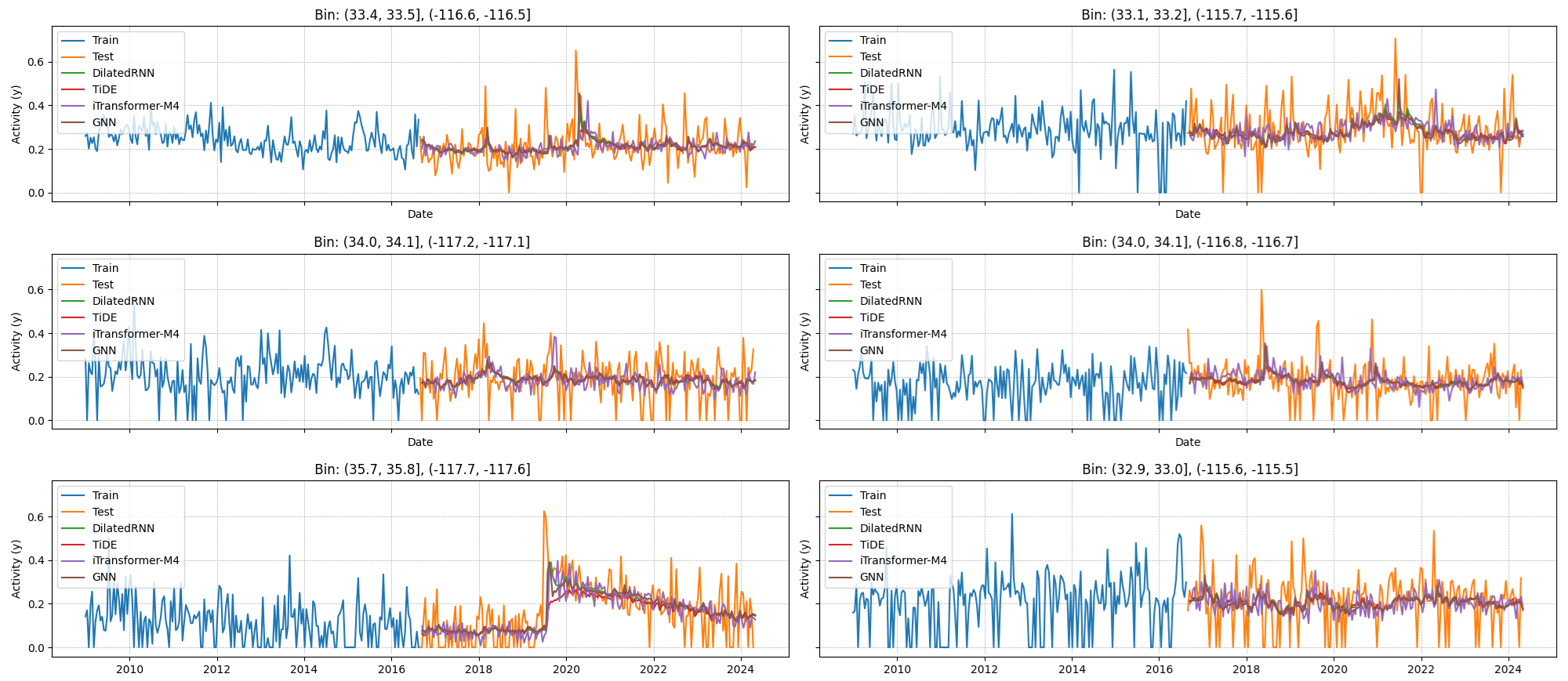}
    \caption{Released energy time series plots for six randomly selected spatial bins, comparing model predictions (GNNCoder one-layer, DilatedRNN, TiDE, iTransformer-M4) against actual observed seismic activities. The GNNCoder and DilatedRNN models effectively capture noticeable spikes in activity, demonstrating their efficacy in short-term earthquake nowcasting and their potential for timely disaster preparedness and response.}
    \label{fig:predicted_6_time_series_plot}
\end{figure}

Figure \ref{fig:predicted_6_time_series_plot} shows the released energy time series plots for six randomly selected spatial bins, providing additional insights into model performance. The plots compare the models' predictions, specifically GNNCoder 1-layer, DilatedRNN, TiDE, and iTransformer-M4, against the actual observed seismic activities over time. Notably, the GNNCoder and DilatedRNN models excel in capturing noticeable spikes in observed activity, illustrating their effectiveness in short-term earthquake nowcasting. This ability to anticipate imminent seismic events is critical for timely disaster preparedness and response, highlighting the practical applicability of these models in real-world scenarios. 

For instance, in the spatial bin (33.4, 33.5), (-116.6, -116.5), the GNN model's predictions closely follow the observed activity trends, while other models like TiDE and iTransformer-M4 show more significant deviations. This pattern is observed across multiple bins, indicating the robustness of the GNNCoder in different spatial contexts.

\subsubsection{Spatial Analysis}
\label{spatial_analysis}
\begin{figure}
    \centering
    \includegraphics[width=0.6\linewidth]{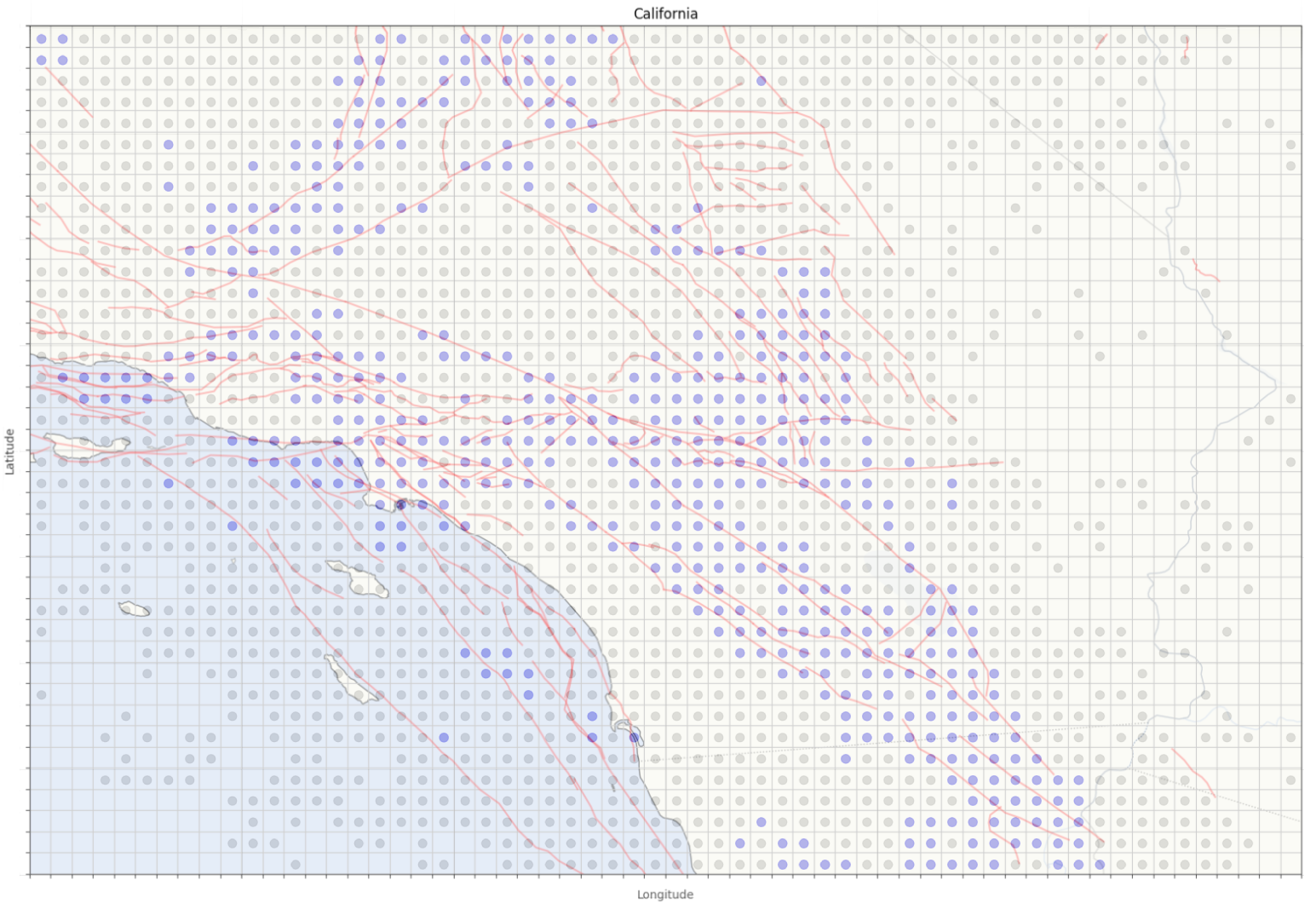}
    \caption{This plot illustrates the spatial bins overlaid on the fault lines to assess the extent to which the fault lines are captured by the bins (graph nodes). It highlights the limitations of the current graph, where some critical fault lines fall outside the spatial bins, impacting the performance of deeper GNN models like the GNNCoder 3-layer model.}
    \label{fig:faults}
\end{figure}

The results presented in table \ref{table:performance_table} demonstrate the importance of considering spatial relationships, where the GNNCoder 1-layer model outperforms the GNNCoder 3-layer model across multiple metrics. This outcome may seem counterintuitive at first, as one might expect a deeper model to capture more intricate patterns and dependencies within the data. However, as shown in Fig \ref{fig:faults}, our approach necessitated the creation of a graph based on spatial bins. These spatial bins serve as nodes in constructing the graph, which unfortunately cannot encompass all parts of fault lines. Consequently, there are bins containing crucial fault information that our graph failed to consider.

The limitation of the graph construction method is particularly detrimental to deeper GNN models, such as the GNNCoder 3-layer model. Deeper models typically rely on the aggregation of information across multiple layers, which can amplify the impact of missing or incomplete data within the graph structure. In this case, the spatial bins that were not included in the graph represent significant gaps in the fault information, hindering the GNNCoder 3-layer model's ability to fully leverage its depth. As a result, the GNNCoder 3-layer model may struggle to effectively capture the underlying patterns of the data, leading to slightly poorer performance compared to the GNNCoder 1-layer model.

In contrast, the GNNCoder 1-layer model, being shallower, is less affected by the incomplete graph representation. Its simpler structure allows it to focus on more immediate, localized relationships within the spatial bins that are included in the graph. This enables the GNNCoder 1-layer model to perform better despite the limitations of the graph construction. Therefore, the results highlight the importance of considering the quality and completeness of the graph structure when designing GNN models for this type of data. A more comprehensive and accurate graph that covers all relevant fault lines might enable deeper GNNCoder models to outperform their shallower counterparts.

Large multivariate transformer-based models, such as PatchTST and iTransformer, utilize channel-independent methods to aggregate input data. In our experiments, the input data for all models consisted of time series from different regions. We expected that these multivariate models would effectively capture the spatial relationships between these time series from our extensive dataset. However, the results reveal that multivariate transformer-based models struggle to accurately capture the latent spatial relationships essential for precise earthquake nowcasting. In contrast, GNNs are particularly adept at understanding spatial dependencies through graph-based analyses that effectively aggregate data from neighboring regions. This localized approach allows GNNs to model the spatial intricacies crucial for accurate earthquake nowcasting.

\subsubsection{Feature Analysis}
Input selection and feature engineering are fundamental components in the development of effective deep-learning models. The identification and utilization of relevant features are pivotal in any field, to the ability to learn and make accurate model predictions. This is particularly true in the field of earthquake nowcasting, where the incorporation of geographical interactions can significantly enhance model performance.

A notable limitation of pre-trained models is the requirement to maintain a consistent input structure during both the pre-training and fine-tuning phases. This restriction hinders the integration of sophisticated domain-specific features that could potentially improve nowcasting performance on earthquake data.

Table \ref{table:performance_table_2} explores the effects of different input configurations, using Multiplicity and Exponential Moving Average (EMA) as inputs, to compare the performance of the top models, GNNCoder, DilatedRNN, and LSTM.

As explained in our papers \citep{2022EA002343, geohazards3020011}, Multiplicity is a critical factor in enhancing the nowcasting accuracy of earthquake nowcasting. Multiplicity refers to the count of earthquake events within a defined spatial-temporal bin that exceeds a specific magnitude threshold, capturing the frequency and intensity of seismic activity and providing a straightforward measure of earthquake occurrence rates. In this study, the magnitude threshold is set at 3.29, and five intervals, ranging from 2 weeks to 260 weeks, are used to calculate Multiplicity. In addition, we employ EMA, which averages the past 5 to 150 samples, further refining the input data.

As shown in Table \ref{table:performance_table_2}, incorporating relevant features enhanced the performance of all three models. The DilatedRNN model with multiplicity and EMA inputs demonstrated the best performance, achieving an MSE of 0.00626, an MAE of 0.0517, and an NNSE of 0.6174. This suggests that the addition of multiplicity and EMA features significantly enhances the DilatedRNN's ability to accurately model sequential patterns in the data.

The GNNCoder with multiplicity and EMA inputs also performed well, with an MSE of 0.00627, an MAE of 0.0517, and an NNSE of 0.6169. Despite its strong performance, the DilatedRNN outperformed the GNNCoder, indicating that the integration of temporal features is particularly beneficial for the memory-based DilatedRNN, enhancing its capacity to capture the temporal complexities of seismic data more effectively.

Additionally, Table \ref{table:performance_table_2} highlights a significant limitation of foundation models, which are constrained to managing only the target time series and cannot accept relevant features.

\begin{table}[ht]
\centering
\setlength{\tabcolsep}{12pt}
\begin{tabular}{l|c|ccccc}
\hline
Model & Input & MSE & MAE  &  NNSE \\
\hline
LSTM & Single feature & 0.00631 & 0.0514 & 0.6156 \\
LSTM  & + Multiplicity & 0.00630 &  0.0506 & 0.6158 \\
DilatedRNN  &  Single feature & 0.00630 & 0.0510 & 0.6159 \\
LSTM  & + Multiplicity + EMA & 0.00629 &  0.0527 &  0.6162 \\
LSTM  & + EMA & 0.00628 & 0.0517  & 0.6164 \\
GNNCoder 1-layer  & + Multiplicity & 0.00628 & 0.0520  & 0.6165 \\
GNNCoder 1-layer  & Single feature & 0.00628 & 0.0522  & 0.6166 \\
GNNCoder 1-layer  &   + Multiplicity + EMA & 0.00627 & 0.0517  & 0.6169 \\
DilatedRNN  &   + Multiplicity & 0.00627 & 0.0517  & 0.6169 \\
GNNCoder 1-layer  & + EMA & 0.00627 & 0.0525  &  0.6172 \\
DilatedRNN  &  + EMA & 0.00627 & 0.0519  & 0.6174\\
DilatedRNN  &   + Multiplicity + EMA & 0.00626 & 0.0517 & 0.6174 \\

\hline
\end{tabular}
\caption{Performance comparison of GNN and DilatedRNN models using various input configurations. The table highlights the impact of incorporating Multiplicity and EMA features on the models' nowcasting accuracy.}
\label{table:performance_table_2}
\end{table}

\section{Conclusion}
\label{conclusion}

This study presents a comprehensive evaluation of foundation models and advanced deep learning architectures for earthquake nowcasting, focusing on Southern California's seismically active region. Our study demonstrates that the selection of appropriate model architectures and pre-training datasets plays a critical role in enhancing nowcasting accuracy for seismic activities.

Our analysis shows that the introduced MultiFoundationQuake model outperforms other models by leveraging the strengths of various foundation models, effectively capturing both temporal and spatial dependencies in seismic data. This model showcases the potential of combining diverse pre-trained models to improve earthquake nowcasting, emphasizing the importance of multi-model integration.

Additionally, the GNNCoder model outperforms other models by effectively capturing spatial relationships and leveraging geographical interactions, resulting in more accurate earthquake nowcasting. Memory-based models like DilatedRNN and LSTM also show strong performance in handling sequential dependencies, though their accuracy could be further improved by incorporating spatial information. The integration of spatial and temporal features is crucial for enhancing nowcasting accuracy.

A notable issue identified in this study is that large multivariate transformer-based models were not successful in capturing the latent spatial relationships between neighboring areas. Future research should focus on enhancing channel-independent methods in multivariate models to better capture subtle spatial dependencies.

In addition, the performance of pre-trained foundation models varied significantly based on the pre-training datasets. Models pre-trained on datasets capturing relevant temporal dynamics, such as iTransformer-TrafficL and PatchTST-Weather, outperformed those pre-trained on less relevant datasets. This finding emphasizes the critical role of selecting appropriate pre-training datasets to improve the models' performance in specific domains. Using pre-training datasets that have the same background as earthquake data may improve earthquake nowcasting by providing more relevant and specific insights. In addition, the results of knowledge transfer learning were not favorable and could be replaced by other methods. For example, the loss function can be designed to prevent the transfer of irrelevant knowledge, where pre-training and fine-tuning datasets are incorporated together.

Future research directions include the development of hybrid models that integrate the strengths of GNNs and RNNs, leveraging both spatial and temporal information to enhance nowcasting capabilities. Improving graph construction methods will also be crucial to better capture the complexities of seismic data. Our study underscores the importance of feature engineering and input selection; incorporating scientific features like Multiplicity and EMA significantly improved model performance. Additionally, integrating more diverse data sources, such as physics equations, could provide a more comprehensive understanding of seismic patterns and further enhance nowcasting accuracy.

Several important geoscience issues should be explored together with the AI topics listed above. These include the spatial extent of the earthquake, which is particularly important as AI models, including spatial links, performed best in this initial study. We will also look at nowcasting over different time periods as in our earlier paper \cite{geohazards3020011}, which looked 4 years into the future from a 2-week time series. We will also try to quantify the origin of the nowcasting accuracy by applying these ideas to simulated ERAS \cite{Rundle2024ERAS} and ETAS earthquakes \cite{Zhuang2012ETAS,Field2017ETAS, Rundle2023ETAS}. We will also explore other geographical regions and different time periods.

This research advances the state of the art in earthquake nowcasting by demonstrating the efficacy of GNNs and pre-trained transformer models. Our improved accuracy and reliability have the potential to enhance disaster response efforts, minimize economic losses, and save lives by providing timely and precise nowcasting of seismic events. This research represents a significant step towards bridging the gap between advanced deep learning methodologies and practical applications in understanding earthquake occurrence and mitigation.

\section*{Acknowledgements}

The University of Virginia authors gratefully acknowledge the partial support of DE-SC0023452: FAIR Surrogate Benchmarks Supporting AI and Simulation Research and the Biocomplexity Institute at the University of Virginia. Research by JBR was supported in part under DoE grant DE-SC0017324 to the University of California, Davis. Portions of this work were carried out at Jet Propulsion Laboratory, California Institute of Technology under contract with NASA. Research by LGL was supported in part by the University of California Irvine Academic Senate award.

%
%I would like to extend my gratitude to OpenAI's ChatGPT for its assistance in the composition and refinement of this paper. The AI capabilities were effective in providing content editing suggestions, grammar checks, and spelling checks.

\bibliographystyle{ACM-Reference-Format}
\bibliography{sample-base.bib}

\end{document}